\documentclass{article}

\usepackage{xspace}
\usepackage{amsfonts}
\usepackage{amssymb}
\usepackage{amsmath}
\usepackage{euscript}
\usepackage{latexsym}
\usepackage{algorithm}
\usepackage{algorithmic}
\usepackage{graphicx}
\usepackage{xcolor}
\usepackage{subfigure}
\usepackage{dsfont}

\newcommand{\BlackBox}{\rule{1.5ex}{1.5ex}}  
\newenvironment{proof}{\par\noindent{\bf Proof\ }}{\hfill\BlackBox\\[2mm]}
 
\newtheorem{theorem}{Theorem}

\renewcommand{\emph}{\textit}
 
\newcommand{\w}{\boldsymbol{w}}
\renewcommand{\t}{\boldsymbol{t}}
\renewcommand{\u}{\boldsymbol{u}}

\renewcommand{\k}{k}

\newcommand{\x}{\boldsymbol{x}}
\newcommand{\y}{\boldsymbol{y}}
\newcommand{\loss}{\ell}

\newcommand{\vtheta}{{\boldsymbol{\theta}}}
\newcommand{\valpha}{{\boldsymbol{\alpha}}}
\newcommand{\vbeta}{{\boldsymbol{\beta}}}
\newcommand{\vgamma}{{\boldsymbol{\gamma}}}

\newcommand{\C}{C}

\newcommand{\one}{\mathbf{1}}
\newcommand{\zero}{\mathbf{0}}

\DeclareMathOperator*{\argmin}{argmin\xspace}

\newcommand{\refprimal}{Optimization Problem (P)\xspace}
\newcommand{\refdual}{Optimization Problem (D)\xspace}

\DeclareMathSymbol{\N}{\mathalpha}{AMSb}{"4E} 
\DeclareMathSymbol{\R}{\mathalpha}{AMSb}{"52}

\def\E{\mathop{\mathbf{E}}}

\begin{document}

\title{A Unifying View of Multiple Kernel Learning}

\author{Marius Kloft \quad Ulrich R\"uckert \quad Peter L. Bartlett\\
University of California, Berkeley, USA\\
\small \{mkloft,rueckert,bartlett\}@cs.berkeley.edu}


\maketitle              

\begin{abstract} 
Recent research on multiple kernel learning has lead to a number of approaches for combining kernels in regularized risk minimization. The proposed approaches include different formulations of objectives and varying regularization strategies. In this paper we present a unifying general optimization criterion for multiple kernel learning and show how existing formulations are subsumed as special cases. We also derive the criterion's dual representation, which is suitable for general smooth optimization algorithms. Finally, we evaluate multiple kernel learning in this framework analytically using a Rademacher complexity bound on the generalization error and empirically in a set of experiments.
\end{abstract} 

\section{Introduction}
Selecting a suitable kernel for a kernel-based \cite{SchSmo02} machine learning task can be a difficult task. From a statistical point of view, the problem of choosing a good kernel is a model selection task. To this end, recent research has come up with a number of \emph{multiple kernel learning} (MKL) \cite{LanCriBarGhaJor04} approaches, which allow for an automated selection of kernels from a predefined family of potential candidates. Typically, MKL approaches come in one of these three different flavors:
\begin{itemize}
\item[(I)] Instead of formulating an optimization criterion with a fixed kernel $k$, one leaves the choice of $k$ as a variable and demands that $k$ is taken from a linear span of base kernels $k := \sum_{i=1}^M \theta_i k_i$. The actual learning procedure then optimizes not only over the parameters of the kernel classifier, but also over the $\vtheta$ subject to the constraint that $\|\vtheta\| \le 1$ for some fixed norm. This approach is taken for instance in \cite{CorMohRos09b} for regression and in \cite{KloBreSonZieLasMue09} for classification.
\item[(II)] A second approach optimizes over all kernel classifiers for each of the $M$ base kernels, but modifies the regularizer to a block norm, that is, a norm of the vector containing the individual kernel norms. This allows to trade-off the contributions of each kernel to the final classifier. This formulation was used for instance in \cite{AflBenBhaNatRamJMLR,Nathetal09}.
\item[(III)] Finally, since it appears to be sensible to have only the best kernels contribute to the final classifier, it makes sense to encourage sparse kernel weights. One way to do so is to extend the second setting with an \emph{elastic net} regularizer, a linear combination of $\ell_1$ and $\ell_2$ regularizers. This approach was recently described in \cite{ryota}.
\end{itemize}
While all of these formulations are based on similar considerations, the individual formulations and used techniques vary considerably. The particular formulations are tailored more towards a specific optimization approach rather than the inherent characteristics. Type (I) approaches, for instance, are generally solved using  a partially dualized wrapper approach, (II) makes use of the fact that the $\ell_\infty$-norm computes a coordinatewise maximum and (III) solves MKL in the primal.
This makes it hard to gain insights into the underpinnings and differences of the individual methods, to design general-purpose optimization procedures for the various criteria and to compare the different techniques empirically.

In this paper, we formulate MKL as an optimization criterion with a dual-block-norm regularizer. By using this specific form of regularization, we can incorporate all the previously mentioned formulations as special cases of a single criterion. We derive a modular dual representation of the criterion, which separates the contribution of the loss function and the regularizer. This allows practitioners to plug in specific (dual) loss functions and to adjust the regularizer in a flexible fashion. We show how the dual optimization problem can be solved using standard smooth optimization techniques, report on experiments on real world data, and compare the various approaches according to their ability to recover sparse kernel weights. On the theoretical side, we give a concentration inequality that bounds the generalization ability of MKL classifiers obtained in the presented framework. The bound is the first known bound to apply to MKL with elastic net regularization and it matches the best previously known bound \cite{CorMohRos10} for the special case of $\ell_1$ and $\ell_2$ regularization.


\section{Generalized MKL}\label{SEC-derivation}

In this section we cast multiple kernel learning in a unified
framework. 
%
Before we go into the details, we need to introduce the general setting and notation.

\subsection{Multiple Kernel Learning}
We begin with reviewing the classical supervised learning setup. Given
a labeled sample $\mathcal{D} = \{(\x_i,y_i)\}_{i=1\ldots,n}$, where
the $\x_i$ lie in some input space $\mathcal{X}$ and $y_i\in\mathcal
Y\subset\mathbb R$, the goal is to find a hypothesis $f \in
\mathcal{H}$,
that generalizes well on new and unseen data. Regularized risk
minimization returns a minimizer $f^*$,
\begin{equation}\label{rrm}
f^* \in \argmin\nolimits_f \,  \text{R}_{\text{emp}}(f) +\lambda\Omega(f)  ,\nonumber
\end{equation}
where $\text{R}_{\text{emp}}(f)=\frac{1}{n}\sum_{i=1}^n
\loss\left(f(\x_i),y_i\right)$ is the empirical risk of hypothesis $f$
w.r.t. a convex loss function $\loss:\mathbb R\times\mathcal
Y\rightarrow\mathbb R$, $\Omega:\mathcal {H} \rightarrow\mathbb R$ is
a regularizer, and $\lambda> 0$ is a trade-off parameter.
We consider linear models of the form
\begin{align}
f_{\w}(\x) = \langle\w,\Phi(\x)\rangle \label{model},
\end{align}
together with a (possibly non-linear) mapping $\Phi:\mathcal{X}\rightarrow\mathcal{H}$
to a Hilbert space $\mathcal{H}$ \cite{MueMikRaeTsuSch01} and constrain the regularization to be of the form
$\Omega(f)=\frac{1}{2}||\w||_2^2$
which allows to kernelize the resulting models and algorithms.
We will later make use of kernel 
functions $k(\x,\x')=\langle \Phi(\x),\Phi(\x') \rangle_{\mathcal{H}}$ to compute inner 
products in $\mathcal{H}$. 

When learning with multiple kernels, we are given $M$ different
feature mappings $\Phi_m:\mathcal{X}\rightarrow\mathcal{H}_m, ~ m=1,\ldots M$, each giving rise
to a reproducing kernel $k_m$ of $\mathcal{H}_m$. There are two main ways to formulate regularized risk minimization with MKL. The first approach introduces a linear kernel mixture $k_\vtheta=\sum_{m=1}^M \theta_m k_m$, $\theta_m\geq0$. With this, one solves 
\begin{align}
  \inf_{\w, \theta} \quad  & \C\sum_{i=1}^n \loss\left(\sum_{m=1}^M\langle \sqrt{\theta_m} \w_m,\Phi(\x_i)\rangle _{\mathcal{H}_m},~y_i\right)
  + \left\| \w_{\vtheta} \right\|_{\mathcal{H}}^2 \label{classicMKL} \\
 \text{s.t.}\quad &  \|\theta\|_q \le 1 \notag , 
\end{align}
with a blockwise weighted target vector $\w_{\vtheta}:=\left(\sqrt{\theta_1}\w_1^\top,..., \sqrt{\theta_M}\w_M^\top\right)^\top$.
Alternatively, one can omit the explicit mixture vector $\theta$ and use block-norm regularization instead. In this case, one optimizes
\begin{align}
  \inf_{\w} \quad  & \C\sum_{i=1}^n \loss\left(\sum_{m=1}^M\langle \w_m,\Phi_m(\x_i)\rangle _{\mathcal{H}_m},~y_i\right)
  + \|\w\|_{2,p}^2 \label{blocknormMKL}
\end{align}
where $||\w||_{2,p}=\left(\sum_{m=1}^M ||\w_m||_{\mathcal H_m}^p\right)^{1/p}$ denotes the $\ell_2/\ell_p$ block norm. One can show that (\ref{classicMKL}) is a special case of (\ref{blocknormMKL}). In particular, one can show that setting the block-norm parameter to $p=\frac{2q}{q+1}$ is equivalent to having kernel mixture regularization with $\|\vtheta\|_q \le 1$ \cite{KloBreSonZieLasMue09}. This also implies that the kernel mixture formulation is strictly less general, because it can not replace block norm regularization for $p > 2$. 
Extending the block norm criterion to also include elastic net \cite{Zou05regularizationand} regularization, we thus choose the following minimization problem as primary object of investigation in this paper:
\\
\\
\textbf{Primal MKL Optimization Problem}
\newcounter{storedequation} 
\let\storedtheequation=\theequation 
\setcounter{storedequation}{\value{equation}} 
\setcounter{equation}{0} 
\renewcommand{\theequation}{P}
\begin{align}
\hspace{-0.3cm}
  \inf_{\w} \quad  \C\sum_{i=1}^n \loss\left(\langle \w,\Phi(\x_i)\rangle _{\mathcal{H}},~y_i\right)
  + \frac{1}{2}||\w||_{2,p}^2 + \frac{\mu}{2}||\w||_2^2 ~,
\end{align}
\setcounter{equation}{\value{storedequation}}  
\renewcommand{\theequation}{\storedtheequation}
\hspace{-0.2cm}where $\Phi=\Phi_1\times\cdots\times\Phi_M$ denotes the cartesian product of the $\Phi_m$'s.
Using the above criterion it is possible to recover block norm regularization by setting $\mu=0$ and the elastic net regularizer by setting $p=1$.

\subsection{Convex MKL in Dual Space}\label{sec:dual}

Optimization problems often have a considerably easier structure when studied in the dual space. 
In this section we derive the dual problem of the generalized MKL approach presented in the previous section.
Let us begin with rewriting \refprimal by expanding the decision values into slack variables as follows
\begin{align}\label{RakoZienConvex}
     \inf_{\w,\t}  & \quad  \C\sum_{i=1}^n \loss\left(t_i,~y_i\right) + \frac{1}{2}||\w||_{2,p}^2 + \frac{\mu}{2}||\w||_2^2\\
     \text{s.t.}  & \quad \forall i: ~  \langle \w,\Phi(\x_i)\rangle_{\mathcal{H}} =t_i . \nonumber
\end{align}
Applying Lagrange's theorem re-incorporates the constraints into the
objective by introducing Lagrangian multipliers
$\valpha\in\mathbb R^n$. \footnote{Note that $\valpha$ is variable over the whole range of $\mathbb R^n$ since it is incorporates an equality constraint.}
The Lagrangian saddle point problem is then given by
\begin{align}\label{lagr}
\sup_{\valpha} ~ \inf_{\w,\t} \quad & \C\sum_{i=1}^n \loss\left(t_i,~y_i\right) + \frac{1}{2}||\w||_{2,p}^2 + \frac{\mu}{2}||\w||_2^2  \\
& -\sum_{i=1}^n\alpha_i\left(\langle \w,\Phi(\x_i)\rangle_{\mathcal{H}} -t_i\right) \nonumber .
\end{align}
Setting the first partial derivatives of the above Lagrangian to zero w.r.t. $\w$
gives  the following KKT optimality condition
\begin{align}
   \forall m: \quad &\w_m ~ =  \left(||\w||_{2,p}^{2-p} ||\w_m||^{p-2}+\mu\right)^{-1} \sum_i  \alpha_i\Phi_m(\x_i) \label{eq:opt_v} ~ .
\end{align}
Inspecting the above equation reveals the representation $\w_m^*\in\text{span}(\Phi_m(\x_1),...,\Phi_m(\x_n))$.
Rearranging the order of terms in the Lagrangian,
\begin{align*}
\sup_{\valpha} \quad &  -\C\sum_{i=1}^n\sup_{\t}\left( -\frac{\alpha_it_i}{C} - \loss\left(t_i,~y_i\right)\right) \nonumber \\
&  -\sup_{\w} \left(\langle \w,\sum_{i=1}^n\alpha_i\Phi(\x_i)\rangle _{\mathcal{H}} - \frac{1}{2}||\w||_{2,p}^2 - \frac{\mu}{2}||\w||_2^2 \right)  ,
\end{align*}
lets us express the Lagrangian in terms of Fenchel-Legendre conjugate functions $h^*(\x)=\sup_{\u} \x^\top\u -h(\u)$ as follows, 
\begin{align}\label{eq444}
\sup_{\valpha} \quad & -\C\sum_{i=1}^n\loss^*\left(-\frac{\alpha_i}{C},~y_i\right) -\left( \frac{1}{2}\left\Vert\sum_{i=1}^n\alpha_i\Phi(\x_i) \right\Vert_{2,p}^2 + \frac{\mu}{2}\left\Vert\sum_{i=1}^n\alpha_i\Phi(\x_i)\right\Vert_2^2\right)^*   ,
\end{align}
thereby removing the dependency of the Lagrangian on $\w$.
The function $\loss^*$ is called \emph{dual loss} in the following.
Recall that the Inf-Convolution 
\cite{Roc70} of two functions $f$ and $g$ is defined by 
$(f\oplus g)(x):=\inf_y f(x-y)+g(y)$ and that 
$(f^*\oplus g^*)(x)=(f+g)^*(x)$, and $(\eta f)^*(x)=\eta f^*(x/\eta)$.
Moreover, we have for the conjugate of the block norm $\left(\frac{1}{2}|| \cdot||^2_{2,p}\right)^*=\frac{1}{2}|| \cdot||^2_{2,p^*}$ \cite{Agarwal:EECS-2008-138} 
where $p^*$ is the conjugate exponent, i.e., $\frac{1}{p}+\frac{1}{p^*}=1$. 
As a consequence, we obtain the following \emph{dual} optimization problem 
\\
\\
\textbf{Dual MKL Optimization Problem} \quad
\newcounter{dualequation} 
\let\dualtheequation=\theequation 
\setcounter{dualequation}{\value{equation}} 
\setcounter{equation}{0} 
\renewcommand{\theequation}{D}
\begin{align}
\sup_{\valpha} \quad &  -\C\sum_{i=1}^n\loss^*\left(-\frac{\alpha_i}{C},~y_i\right) -\left( \frac{1}{2}\left\Vert \cdot \right\Vert_{2,p*}^2 \oplus \frac{1}{2\mu}\left\Vert  \cdot\right\Vert_2^2\right )\left(\sum_{i=1}^n\alpha_i\Phi(\x_i)\right)   .
\end{align}
\setcounter{equation}{\value{dualequation}}  
\renewcommand{\theequation}{\dualtheequation}Note that the supremum is also a maximum, if the loss function is continuous. The function $f\oplus\frac{1}{2\mu}||\cdot||^2$ is the so-called \emph{Morea-Yosida Approximate} \cite{morea} and has been studied extensively both theoretically and algorithmically for its favorable regularization properties. It can ``smoothen'' an optimization problem---even if it  is  initially non-differentiable---and it increases the condition number of the Hessian for twice differentiable problems. 

The above dual generalizes multiple kernel learning to arbitrary convex loss functions and regularizers. 
Due to the mathematically clean separation of the loss and the regularization term---each loss term solely depends on a single real valued variable---we can
immediately recover the corresponding dual for a specific choice of a loss/regularizer pair $(\loss,||\cdot||_{2,p})$ by computing the pair of
conjugates $(\loss^*,||\cdot||_{2,p^*})$.

\subsection{Obtaining Kernel Weights}\label{SEC-applications}
While formalizing multiple kernel learning with block-norm regularization offers a number of conceptual and analytical advantages, it requires an additional step in practical applications. The reason for this is that the block-norm regularized dual optimization criterion does not include explicit kernel weights. Instead, this information is contained only implicitly in the optimal kernel classifier parameters, as output by the optimizer. This is a problem, for instance if one wishes to apply the induced classifier on new test instances. Here we need the kernel weights to form the final kernel used for the actual prediction. To recover the underlying kernel weights, one essentially needs to identify which kernel contributed to which degree for the selection of the optimal dual solution. Depending on the actual parameterization of the primal criterion, this can be done in various ways.

We start by reconsidering the KKT optimality condition given by Eq.~\eqref{eq:opt_v} and
observe that the first term on the right hand side,
\begin{equation}\label{eq:opttheta}
  \theta_m:=  \left(||\w||_{2,p}^{2-p} ||\w_m||^{p-2}+\mu\right)^{-1} .
\end{equation}
introduces a scaling of the feature maps. With this notation, it is easy to see from Eq.~\eqref{eq:opt_v} that our model given by Eq.~\eqref{model} extends to 
$$
  f_{\w}(\x) = \sum_{m=1}^M\sum_{i=1}^n\alpha_i\theta_m\k_m(\x_i,\x) .
$$
In order to express the above model solely in terms of dual varables we have to compute $\vtheta$ in terms of $\valpha$.

In the following we focus on two cases. First, we consider $\ell_p$ block norm regularization for arbritrary $1< p<\infty$ while switching the elastic net off by setting the parameter $\mu=0$.
Then, from Eq.~\eqref{eq:opt_v} we obtain 
$$ ||\w_m||=||\w||_{2,p}^{\frac{p-2}{p-1}} \left\Vert\sum_{i=1}^n \alpha_i \Phi_m(\x_i)\right\Vert_{\mathcal H_m}^{\frac{1}{p-1}}  \quad \text{where} \quad \w_m = \theta_m \sum_i  \alpha_i\Phi_m(\x_i)  .$$
Resubstitution into \eqref{eq:opttheta} leads to the proportionality
\begin{equation}\label{eq:opt1}
  \exists~ c>0 ~~ \forall ~m:\quad  \theta_m ~  =  ~ c\left(\left\Vert \sum_{i=1}^n\alpha_i\Phi_m(\x_i)\right\Vert_{\mathcal H_m}\right)^{\frac{2-p}{p-1}} .
\end{equation}
Note that, in the case of classification, we only need to compute $\vtheta$ up to a positive multiplicative constant.

For the second case, let us now consider the elastic net regularizer, i.e., $p=1+\epsilon$ with $\epsilon\approx 0$ and $\mu>0$. Then, the optimality condition given by Eq.~\eqref{eq:opt_v} translates to
$$ \w_m = \theta_m \sum_i  \alpha_i\Phi_m(\x_i)  \quad \text{where} \quad \theta_m =\left(\left(\sum_{m'=1}^M ||\w_{m'}||^{1+\epsilon}_{\mathcal H_{m'}}\right)^{1-\epsilon} ||\w_m||_{\mathcal H_m}^{\epsilon-1} +\mu\right)^{-1} .$$
Inserting the left hand side expression for $||\w_m||_{\mathcal H_{m}}$ into the right hand side leads to the non-linear system of equalities
\begin{align}\label{eq:opt2}
  \forall ~ m: ~ ~ \mu\theta_m||K_m||^{1-\epsilon} + \theta_m^{\epsilon}\left(\sum_{m'=1}^M\theta_{m'}^{1+\epsilon}||K_{m'}||^{1+\epsilon}\right)^{1-\epsilon} = ||K_m||^{1-\epsilon}, 
\end{align}
where we employ the notation $||K_m||:=\left\Vert\sum_{i=1}^n \alpha_i \Phi_m(\x_i)\right\Vert_{\mathcal H_m}$.
In our experiments we solve the above conditions numerically using $\epsilon\approx 0$.
The optimal mixing coefficients $\theta_m$ can now be computed 
solely from the dual $\valpha$ variables by means of Eq.~\eqref{eq:opt1}~and~\eqref{eq:opt2}, and by the kernel matrices $K_m$ using the identity 
$$\forall m=1,\cdots,M: ~ ~ ||K_m||=\sqrt{\valpha K_m \valpha}.$$
This enables optimization in the dual space as discussed in the next section.

\section{Optimization Strategies}\label{SEC-optimization}

In this section we describe how one can solve the dual optimization problem using an efficient quasi-Newton method. For our experiments, we use the hinge loss $l(x) =\max(0,1-x)$, but the discussion also applies to most other convex loss functions.
We first note that the dual loss of the hinge loss is $\loss^*(t,y)=\frac{t}{y}$ if
$-1\leq\frac{t}{y}\leq 0$ and $\infty$ elsewise \cite{RifLip07}.
Hence, for each $i$ the term $\loss^*\left(-\frac{\alpha_i}{\C},~y_i\right)$ of
the generalized dual, i.e., \refdual, translates to 
$-\frac{\alpha_i}{Cy_i}$, provided that $0\leq \frac{\alpha_i}{y_i}\leq C$. Employing a variable substitution of the form
$\alpha^{\text{new}}_i=\frac{\alpha_i}{y_i}$, 
the dual problem (D) becomes
\begin{align*}
\sup_{\valpha:~\zero\leq\valpha\leq\one} \quad &  \one^\top\valpha -\left( \frac{1}{2}\left\Vert \cdot \right\Vert_{2,p*}^2 \oplus \frac{1}{2\mu}\left\Vert  \cdot\right\Vert_2^2\right )\left(\sum_{i=1}^n\alpha_iy_i\Phi(\x_i)\right)   ,
\end{align*}
and by definition of the Inf-convolution,
\begin{align}
\sup_{\valpha,\vbeta:~\zero\leq\valpha\leq\one} \quad &  \one^\top\valpha - \frac{1}{2}\left\Vert \sum_{i=1}^n\alpha_iy_i\Phi(\x_i)-\vbeta \right\Vert_{2,p*}^2 - \frac{1}{2\mu}\left\Vert \vbeta\right\Vert_2^2   . \label{eq555}
\end{align}
We note that the representer theorem \cite{SchSmo02} is valid for the above problem, and hence
the solution of \eqref{eq555} can be expressed in terms of kernel functions, i.e.,
$\vbeta_m=\sum_{i=1}^n \gamma_i k_m(x_i,\cdot)$  for certain real coefficients $\vgamma\in\mathbb R^n$ uniformly for all $m$,  hence $\vbeta=\sum_{i=1}^n \gamma_i\Phi(x_i)$. 
Thus, Eq.~\ref{eq555} has a representation of the form
\begin{align*}
  \sup_{\valpha,\vgamma:~\zero\leq\valpha\leq\one} \quad &  \one^\top\valpha - \frac{1}{2}\left\Vert \sum_{i=1}^n(\alpha_iy_i-\gamma_i)\Phi(\x_i) \right\Vert_{2,p*}^2 -  \frac{1}{2\mu}\vgamma \top K \vgamma   .
\end{align*}
The above expression can be written\footnote{We employ the notation $s=(s_1,\ldots,s_M)^\top=(s_m)_{m=1}^M$ for $s\in \mathbb R^M$.} in terms of kernel matrices as follows,
\\
\\
\textbf{Hinge Loss Dual Optimization Problem} \quad
\newcounter{hingedualequation} 
\let\hingedualtheequation=\theequation 
\setcounter{hingedualequation}{\value{equation}} 
\setcounter{equation}{0} 
\renewcommand{\theequation}{D'}
\begin{align}
  \sup_{\valpha,\vgamma:~\zero\leq\valpha\leq\one} \quad &  \one^\top\valpha - \frac{1}{2}\left\Vert\left(\sqrt{(\valpha\circ\y-\vgamma)^\top K_m (\valpha\circ\y-\vgamma)}\right)_{m=1}^M \right\Vert_{p*}^2 - \frac{1}{2\mu}\vgamma \top K \vgamma     , \label{eqoptim}
\end{align}\setcounter{equation}{\value{hingedualequation}}  
\renewcommand{\theequation}{\hingedualtheequation}
\hspace{-0.2cm}where we denote by $\x\circ\y$ the elementwise multiplication of two vectors and use the shorthand $K=\sum_{m=1}^M K_m$.

\section{Theoretical Results}\label{SEC-theory}

In this section we give two uniform convergence bounds for the generalization error of the multiple kernel learning formulation presented in Section \ref{SEC-derivation}. The results are based on the established theory on Rademacher complexities. Let $\sigma_1, \ldots, \sigma_n$ be a set of independent Rademacher variables, which obtain the values -1 or +1 with the same probability 0.5. and let $\mathcal{C}$ be some space of classifiers $c: \mathcal{X} \rightarrow \R$. Then, the \emph{Rademacher complexity} of $\mathcal{C}$ is given by
\begin{align}
\mathcal{R}_{\mathcal{C}} & := \E \left[ \sup_{c \in \mathcal{C}} \frac{1}{n} \sum_{i=1}^n \sigma_i c(x_i) \right] \notag~.
\end{align}
If the Rademacher complexity of a class of classifiers is known, it can be used to bound the generalization error. We give one result here, and refer to the literature \cite{BarMen02} for further results on Rademacher penalization.
\begin{theorem}
 Assume the loss $\loss: \R \rightarrow \R$ has $\loss(0)=0$, is Lipschitz with constant $L$ and $\loss(x) \le 1$ for all $x$.  Then, the following holds with probability larger than $1-\delta$ for all classifiers $c \in \mathcal{C}$: 
\begin{align}
\E[\loss(yc(x))] \le \frac{1}{n} \sum_{i=1}^n \loss( y_i c(x_i)) + 2L \mathcal{R}_{\mathcal{C}} +  \sqrt{\frac{8 \ln \frac{2}{\delta}}{n}} \label{RademacherBound}~.
\end{align}
\end{theorem}
We will now give an upper bound for the Rademacher complexity of the block-norm regularized linear learning approach described above. More precisely, for $1 \le i \le M$ let $\|w\|_{\star i} := \sqrt{k_i(w,w)}$ denote the norm induced by kernel $k_i$ and for $x \in \R^p$, $p,q \ge1$ and $C_1, C_2 \ge 0$ with $C_1 + C_2=1$ define
\begin{align}
\|x\|_O := C_1 \|x\|_p + C_2 \|x\|_q \notag .
\end{align}
We now give a bound for the following class of linear classifiers:
\begin{align}
\mathcal{C}_{\star} &:= \left\{  c: \left( \begin{array}{c}  \Phi_1(x) \\ \vdots \\  \Phi_M(x) \end{array} \right) \mapsto \left( \begin{array}{c}  w_1 \\ \vdots \\  w_M \end{array} \right)^T  \left( \begin{array}{c}  \Phi_1(x) \\ \vdots \\  \Phi_M(x) \end{array} \right)  \bigg| \left\|\left( \begin{array}{c}  \|w_1\|_{\star 1} \\ \vdots \\  \|w_M\|_{\star M} \end{array} \right)\right\|_O \le 1 \right\} \notag .
\end{align} 
\begin{theorem}\label{MainTheorem}
Assume the kernels are normalized, i.e. $k_i(x,x) = \|x\|_{\star i}^2 \le 1$ for all $x \in \mathcal{X}$ and all $1 \le i \le M$. Then, the Rademacher complexity of the class $\mathcal{C}_{\star}$ of linear classifiers with block norm regularization is upper-bounded as follows:
\begin{align}
\mathcal{R}_{\mathcal{C}_{\star}} \le  \frac{M}{C_1 M^{\frac{1}{p}} + C_2 M^{\frac{1}{q}}}  \left(\sqrt{\frac{2\ln M}{n}} + \sqrt{\frac{1}{n}}\right) \label{MainBound}.
\end{align}
For the special case with $p \ge 2$ and $q \ge 2$, the bound can be improved as follows:
\begin{align}
\mathcal{R}_{\mathcal{C}_{\star}} \le  \frac{M}{C_1 M^{\frac{1}{p}} + C_2 M^{\frac{1}{q}}}  \sqrt{\frac{1}{n}} \label{MainResSpecialCase}~.
\end{align}
\end{theorem}
It is instructive to compare this result to some of the existing MKL bounds in the literature. For instance, the main result in \cite{CorMohRos10} bounds the Rademacher complexity of the $\ell_1$-norm regularizer with a $O(\sqrt{\ln M / n})$ term. We get the same result by setting $C_1=1, C_2=0$ and $p=1$. For the $\ell_2$-norm regularized setting, we can set $C_1=1, C_2=0$ and $p=\frac{4}{3}$ (because the kernel weight formulation with $\ell_2$ norm corresponds to the block-norm representation with $p=\frac{4}{3}$) to recover their $O(M^{\frac{1}{4}}/\sqrt{n})$ bound. Finally, it is interesting to see how changing the $C_1$ parameter influences the generalization capacity of the elastic net regularizer ($p=1, q=2$). For $C_1=1$, we essentially recover the $\ell_1$ regularization penalty, but as $C_1$ approaches 0, the bound includes an additional $O(\sqrt{M})$ term. This shows how the capacity of the elastic net regularizer increases towards the $\ell_2$ setting with decreasing sparsity.     
\begin{proof}[of Theorem \ref{MainTheorem}]
Using the notation $w:=(w_1, \ldots, w_M)^T$ and $\|w\|_B := \|(\|w_1\|_{\star 1}, \ldots, \|w_M\|_{\star M})^T \|_O$ it is easy to see that
\begin{align}
\E \left[\sup_{c \in \mathcal{C}_{\star}} \frac{1}{n}\sum_{i=1}^n \sigma_i y_i c(x_i) \right] &= 
\E\left[ \sup_{\|w\|_B \le 1} \left\{ \left( \begin{array}{c}  w_1 \\ \vdots \\  w_M \end{array} \right)^T \left( \begin{array}{c} \frac{1}{n}\sum_{i=1}^n\sigma_i\Phi_1(x_i) \\ \vdots \\  \frac{1}{n}\sum_{i=1}^n\sigma_i \Phi_M(x_i)\end{array} \right) \right\} \right] \notag \\
&= \E\left[ \left\| \left( \begin{array}{c} \|  \frac{1}{n}\sum_{i=1}^n\sigma_i\Phi_1(x_i)\|_{\star1} \\ \vdots \\ \|  \frac{1}{n}\sum_{i=1}^n\sigma_i \Phi_M(x_i)\|_{\star M} \end{array} \right) \right\|_O^* \right] \notag ,
\end{align}
where $\|x\|^* := \sup_{z} \{ z^Tx | \|z\| \le 1\}$ denotes the dual norm of $\|.\|$ and we use the fact that $\|w\|_B^*= \|(\|w_1\|^*_{\star 1}, \ldots, \|w_M\|^*_{\star M})^T \|_O^*$ \cite{Agarwal:EECS-2008-138}, and that $\|.\|_{\star i}^* = \|.\|_{\star i}$.
We will show that this quantity is upper bounded by
\begin{align}
 \frac{M}{C_1 M^{\frac{1}{p}} + C_2 M^{\frac{1}{q}}}  \left(\sqrt{\frac{2\ln M}{n}} + \sqrt{\frac{1}{n}}\right) \label{MainRes}.
\end{align}
As a first step we prove that for any $x \in \R^M$
\begin{align}
\|x\|_O^* \le \frac{M}{C_1 M^{\frac{1}{p}} + C_2 M^{\frac{1}{q}}}\|x\|_\infty \label{NormStep}~.
\end{align}
For any $a \ge 1$ we can apply H\"older's inequality to the dot product of $x \in \R^M$ and $\mathds{1}_M := (1, \ldots, 1)^T$ and obtain
$\|x\|_1  \le \|\mathds{1}_M\|_{\frac{a}{a-1}} \cdot \|x\|_a = M^{\frac{a-1}{a}} \|x\|_a $.
Since $C_1+C_2=1$, we can apply this twice on the two components of $\|.\|_O$ to get a lower bound for $\|x\|_O$,
$$
(C_1M^{\frac{1-p}{p}} + C_2M^{\frac{1-q}{q}})\|x\|_1 ~ \le ~C_1\|x\|_p + C_2\|x\|_q ~= ~\|x\|_O.
$$
In other words, for every $x \in \R^M$ with $\|x\|_O \le 1$ it holds that
\begin{align}
\|x\|_1\le 1/\left(C_1M^{\frac{1-p}{p}} + C_2M^{\frac{1-q}{q}}\right) = M/\left(C_1M^{\frac{1}{p}} + C_2M^{\frac{1}{q}}\right) \notag.
\end{align}
Thus,
\begin{align}
\left\{z^Tx | \|x\|_O \le 1\right\} \subseteq \left\{z^Tx \Big| \|x\|_1 \le  \frac{M}{C_1M^{\frac{1}{p}} + C_2M^{\frac{1}{q}}}\right\} \label{dualbound1}.
\end{align}
This means we can bound the dual norm $\|.\|_O^*$ of $\|.\|_O$ as follows:
\begin{align}
\|x\|_O^* &= \sup_z \{z^Tx | \|z\|_O \le 1\} \notag \\
&\le \sup_z \left\{z^Tx \Big| \|z\|_1 \le  \frac{M}{C_1M^{\frac{1}{p}} + C_2M^{\frac{1}{q}}}\right\} \notag \\
&=  \frac{M}{C_1M^{\frac{1}{p}} + C_2M^{\frac{1}{q}}}\|x\|_\infty \label{dualbound2} ~.
\end{align}
This accounts for the first factor in (\ref{MainRes}). For the second factor, we show that
\begin{align}
\E\left[ \left\| \left( \begin{array}{c} \|  \frac{1}{n}\sum_{i=1}^n\sigma_i\Phi_1(x_i)\|_{\star1} \\ \vdots \\ \|  \frac{1}{n}\sum_{i=1}^n\sigma_i \Phi_M(x_i)\|_{\star M} \end{array} \right) \right\|_\infty \right] &\le \sqrt{\frac{2\ln M}{n}} + \sqrt{\frac{1}{n}}\label{MainRes2} ~.
\end{align}
To do so, define
$$
V_k := \left\|\frac{1}{n} \sum_{i=1}^n \sigma_i \Phi_k(x_i)\right\|_{\star k}^2 
 = \frac{1}{n^2}\sum_{i=1}^n\sum_{j=1}^n \sigma_i \sigma_j k_k(x_i, x_j) ~.
$$
By the independence of the Rademacher variables it follows for all $k \le M$,
\begin{align}
\E \left[ V_k \right] 
= \frac{1}{n^2}\sum_{i=1}^n \E\left[k_k(x_i, x_i)\right]  \le \frac{1}{n} \label{ExpectationBound}.
\end{align}

In the next step we use a martingale argument to find an upper bound for $\sup_k [W_k]$ where $W_k := \sqrt{V_k} - \E[\sqrt{V_k}]$. For ease of notation, we write $\E_{(r)} [X]$ to denote the conditional expectation $\E[X|(x_1,\sigma_1), \ldots (x_r, \sigma_r)]$. We define the following martingale:
\begin{align}
Z_k^{(r)} &:= \E_{(r)}[\sqrt{V_k}]- \E_{(r-1)}[\sqrt{V_k}] \notag \\
&= \E_{(r)}\left[\left\|\frac{1}{n}\sum_{i=1}^n \sigma_i \Phi_k(x_i)\right\|_{\star k}\right]- \E_{(r-1)}\left[\left\|\frac{1}{n}\sum_{i=1}^n \sigma_i \Phi_k(x_i)\right\|_{\star k}\right]. \label{martingale1}
\end{align}
The range of each random variable $Z_k^{(r)}$ is at most $\frac{2}{n}$. This is because switching the sign of $\sigma_r$ changes only one summand in the sum from $-\Phi_k(x_r)$ to $+\Phi_k(x_r)$. Thus, the random variable changes by at most $\|\frac{2}{n} \Phi_k(x_r)\|_{\star k} \le \frac{2}{n} k_k(x_r,x_r) \le 1$.
Hence, we can apply Hoeffding's inequality, 
$  \E_{(r-1)} \left[e^{sZ^{(k)}_r}\right]\le e^{\frac{1}{2n^2}s^2}  .$

This allows us to bound the expectation of $\sup_k W_k$ as follows:
\begin{align}
\E[\sup_k W_k] &= \E \left[\frac{1}{s} \ln \sup_k  e^{sW_k}\right] \notag \\
&\le \E\left[\frac{1}{s} \ln \sum_{k=1}^M \exp \left[ s\sum_{r=1}^n Z_k^{(r)}\right] \right] \notag \\
&\le \frac{1}{s} \ln \sum_{k=1}^M  \E\left[e^{ s \sum_{r=1}^{n-1} Z_k^{(r)}}\E_{(n-1)}\left[e^{s Z_k^{(n)}}\right] \right] \notag \\
&\le \frac{1}{s} \ln \sum_{k=1}^M  \left(e^{\frac{1}{2n^2}s^2}\right)^n\notag\\
&= \frac{\ln M}{s} +\frac{s}{2n}\notag ~,
\end{align}
where we $n$ times applied Hoeffding's inequality.
Setting $s=\sqrt{2n \ln M}$ yields:
\begin{align}
\E[\sup_k W_k] &\le \sqrt{\frac{2\ln M}{n}} \label{VarBound} ~.
\end{align}
Now, we can combine (\ref{ExpectationBound}) and (\ref{VarBound}):
$$
\E\left[\sup_k \sqrt{V_k}\right] \le\E\left[ \sup_k W_k+ \sqrt{\E[V_k]}\right] \le \sqrt{\frac{2\ln M}{n}} + \sqrt{\frac{1}{n}} ~.
$$
This concludes the proof of (\ref{MainRes2}) and therewith (\ref{MainBound}).

The special case (\ref{MainResSpecialCase}) for $p,q \ge 2$ is similar.
 As a first step, we modify (\ref{NormStep}) to deal with the $\ell_2$-norm rather than the $\ell_\infty$-norm:
\begin{align}
\|x\|_O^* &\le \frac{\sqrt{M}}{C_1 M^{\frac{1}{p}} + C_2 M^{\frac{1}{q}}}\|x\|_2 \label{secondnormbound}~.
\end{align}
To see this, observe that for any $x \in \R^M$ and any $a \ge 2$ H\"older's inequality gives
$\|x\|_2 \le M^{\frac{a-2}{2a}} \|x\|_a$. 
Applying this to the two components of $\|.\|_O$ we have: 
\begin{align}
(C_1M^{\frac{2-p}{2p}} + C_2M^{\frac{2-q}{2q}})\|x\|_2  
& \le C_1\|x\|_p + C_2\|x\|_q = \|x\|_O \notag ~.
\end{align}
In other words, for every $x \in \R^M$ with $\|x\|_O \le 1$ it holds that
\begin{align}
\|x\|_2\le 1/\left(C_1M^{\frac{2-p}{2p}} + C_2M^{\frac{2-q}{2q}}\right) = {\sqrt{M}}/\left(C_1M^{\frac{1}{p}} + C_2M^{\frac{1}{q}}\right) \notag.
\end{align}
Following the same arguments as in (\ref{dualbound1}) and (\ref{dualbound2}) we obtain (\ref{secondnormbound}). To finish the proof it now suffices to show that
\begin{align}
\E\left[ \left\| \left( \begin{array}{c} \|  \frac{1}{n}\sum_{i=1}^n\sigma_i\Phi_1(x_i)\|_{\star1} \\ \vdots \\ \|  \frac{1}{n}\sum_{i=1}^n\sigma_i \Phi_M(x_i)\|_{\star M} \end{array} \right) \right\|_2 \right] &\le \sqrt{\frac{M}{n}} \notag~.
\end{align}
This is can be seen by a straightforward application of (\ref{ExpectationBound}):
\begin{align}
\E \left[ \sqrt{\sum_{k=1}^M  \left\|  \frac{1}{n}\sum_{i=1}^n\sigma_i\Phi_k(x_i)\right\|_{\star k}^2} \right] &\le \sqrt{ \E \left[\sum_{k=1}^M  V_k \right] }\notag 
\le \sqrt{\sum_{i=1}^M \frac{1}{n} }
= \sqrt{\frac{M}{n}} \notag~.
\end{align}

\end{proof}

\section{Empirical Results}\label{SEC-experiments}

In this section we evaluate the proposed method on artifical and real data sets.
We chose the limited memory quasi-Newton software L-BFGS-B  \cite{ZhuByrLuNoc97} to solve \eqref{eqoptim}.
L-BFGS-B approximates the Hessian matrix based on the last $t$ gradients, where $t$ is a  parameter to be chosen by the user.

\subsection{Experiments with Sparse and Non-Sparse Kernel Sets}\label{sectoy1}
\begin{figure}[t]
  \centering
  \includegraphics[width=0.8\textwidth]{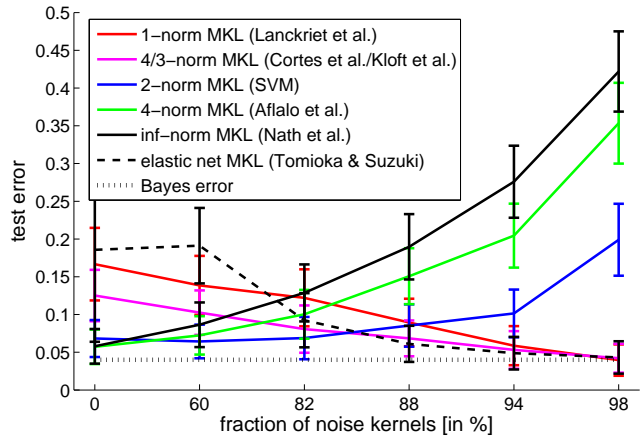}
  \caption{  \small
  Empirical results of the artificial experiment for varying true underlying data sparsity.}
  \label{toy-gap50}
\end{figure}

The goal of this section is to study the relationship of the level of sparsity
of the true underlying function to the chosen block norm or elastic net MKL model.
Apart from investigating which parameter choice leads to optimal results,
we are also interested in the effects of suboptimal choices of $p$.
To this aim we constructed several artificial data sets in which we vary 
the degree of sparsity in the true kernel mixture coefficients.
We go from having all weight focussed on a single kernel (the highest level of sparsity)
to uniform weights (the least sparse scenario possible) in several steps.
We then study the statistical performance of $\ell_p$-block-norm MKL for different
values of $p$ that cover the entire range $[0,\infty]$.
We follow the experimental setup of \cite{Kloft:EECS-2010-21} but compute classification models for 
 $p=1,4/3,2,4,\infty$ block-norm MKL and $\mu=10$ elastic net MKL. 
The results are shown in Fig.~\ref{toy-gap50} and compared to the Bayes error that is computed analytically from the underlying probability model.

Unsurprisingly, $\ell_1$ performs best in 
the sparse scenario, where only a single kernel carries the whole discriminative information of the learning problem. 
In contrast, the $\ell_\infty$-norm MKL performs best when all kernels are equally informative. Both MKL variants reach the Bayes error in their respective scenarios.
The elastic net MKL performs comparable to $\ell_1$-block-norm MKL. The non-sparse  $\ell_{4/3}$-norm MKL and the unweighted-sum kernel SVM perform best in the balanced scenarios, i.e., when the noise level is ranging in the interval 60\%-92\%. 
The non-sparse $\ell_4$-norm MKL of \cite{AflBenBhaNatRamJMLR} performs only well in the most non-sparse scenarios.
Intuitively, the non-sparse $\ell_{4/3}$-norm MKL of \cite{CorMohRos09b,KloBreSonZieLasMue09}  is the most robust MKL variant, achieving an test error of less than $0.1\%$ in all scenarios.
The sparse $\ell_1$-norm MKL performs worst when the noise level is  less than $82\%$.
It is worth mentioning that when considering the most challenging model/scenario combination, that is $\ell_\infty$-norm in the sparse and $\ell_1$-norm in the uniformly non-sparse scenario,
the $\ell_1$-norm MKL performs much more robust than its $\ell_\infty$ counterpart. 
However, as witnessed in the following sections, this does not prevent $\ell_\infty$ norm MKL from performing very well in practice.
In summary, we conclude that by tuning the sparsity parameter $p$ for each experiment,   
block norm MKL achieves a low test error across all scenarios.

\subsection{Gene Start Recognition}\label{sec:tss}

This experiment aims 
at detecting transcription start sites (TSS) of
RNA Polymerase II binding genes in genomic DNA sequences.
Accurate detection of the
transcription start site is crucial to identify genes and their
promoter regions and can be regarded as a first step in
deciphering the key regulatory elements in the promoter region that
determine transcription.

Many  detectors thereby rely on a combination of 
feature sets which makes the learning task appealing for MKL. 
For our experiments we use the data set from \cite{SonZieRae06} 
and we employ five different kernels representing the TSS signal 
(weighted degree with shift), the  promoter (spectrum), 
the 1st exon (spectrum), angles (linear), 
and energies (linear).
The kernel matrices are  normalized such that each feature vector has unit norm in Hilbert space.
We reserve 500 and 500 randomly drawn instances for holdout and 
test sets, respectively, and use 1,000 as the training pool from which 250 elemental training sets are drawn.
Table~\ref{tab:bio} shows the area under the ROC curve (AUC)
averaged over 250 repetitions of the experiment. Thereby $1$ and $\infty$ block norms are approximated
by $64/63$ and $64$ norms, respectively. For the elastic net we use an $\ell_{1.05}$-block-norm penalty.

\begin{table}
\begin{center}
\caption{Results for the bioinformatics experiment.}
\label{tab:bio}
\medskip
\begin{tabular}{l|l}
            & AUC $\pm$ stderr     \\\hline\hline
$\mu=0.01$ elastic net     & $\mathbf{85.80} \pm \mathbf{0.21}$ \\
$\mu=0.1$ elastic net    &   $85.66 \pm 0.15$ \\
$\mu=1$ elastic net    &    $83.75 \pm 0.14$  \\
$\mu=10$ elastic net    &   $84.56 \pm 0.13$  \\
$\mu=100$ elastic net    &  $84.07 \pm 0.18$  \\
\hline
$1$-block-norm MKL & $ 84.83 \pm 0.12$ \\
${4/3}$-block-norm MKL & $85.66 \pm 0.12$  \\
$2$-block-norm MKL     &  $85.25 \pm 0.11$  \\
$4$-block-norm    MKL   &   $85.28 \pm 0.10$  \\
$\infty$-block-norm  MKL    &  $\mathbf{87.67} \pm \mathbf{0.09}$ \\
\end{tabular}
\end{center}
\end{table}

The results vary greatly between the MKL models.
The elastic net model gives the best prediction for $\mu=0.01$ by essentially approximating the $\ell_{1.05}$-block-norm MKL. 
Out of the block norm MKLs the classical $\ell_1$-norm MKL has the worst prediction accuracy and
is even outperformed by an unweighted-sum kernel SVM (i.e., $p=2$ norm MKL). 
In accordance with previous experiments in \cite{KloBreSonZieLasMue09} 
the $p=4/3$-block-norm has the highest prediction accuracy of the models within the parameter range $p\in [1,2]$.
Surprisingly, this superior performance can even be improved considerably by the recent $\ell_\infty$-block-norm MKL of \cite{Nathetal09}.
This is remarkable, and of significance for the application domain: the method using the
unweighted sum of kernels \cite{SonZieRae06} has recently been
confirmed to be the leading in a comparison of 19 state-of-the-art
promoter prediction programs \cite{AbeelPS09}, and our experiments
suggest that its accuracy can be further improved by $\ell_\infty$ MKL.

\subsection{Network Intrusion Detection}\label{sec-exectime}

For the intrusion detection experiments we use the data set described in \cite{KloNakBre09} consisting of HTTP traffic 
recorded at Fraunhofer Institute FIRST Berlin. 
The unsanitized data contains 500 normal HTTP requests drawn
randomly from incoming traffic recorded over two months.
Malicious traffic is generated using the Metasploit framework \cite{MayMooCerRosBea07} and consists of
 30 instances of 10 real attack classes from recent
exploits, including buffer overflows and PHP vulnerabilities. 
Every attack is recorded in different variants using virtual network 
environments and decoy HTTP servers.

We deploy 10 spectrum kernels \cite{LesEskNob02,ShaCri04} for 
$1,2,\ldots,10$-gram feature representations. 
All data points are 
normalized to unit norm in feature space to avoid 
dependencies on the HTTP request length.
We randomly split the normal data into 100 training, 200 validation and 250 test 
examples. We report on average areas under the ROC curve in the false-positive 
interval $[0,0.1]$ (AUC$_{[0,0.1]}$) over 100 repetitions with distinct
training, holdout, and test sets.

\begin{table}
\begin{center}
\caption{Results for the intrusion detection experiment.}
\label{tab:ids}
\medskip
\begin{tabular}{l|l}
            & AUC$_{0.1}$ $\pm$ stderr     \\\hline\hline
$\mu=0.01$ elastic net    & $99.36 \pm 0.14$ \\
$\mu=0.1$ elastic net    & $\mathbf{99.46} \pm \mathbf{0.13}$ \\
$\mu=1$ elastic net    &  $99.38 \pm 0.12$ \\
$\mu=10$ elastic net    &  $99.43 \pm 0.11$ \\
$\mu=100$ elastic net    &  $ 99.34 \pm 0.13$ \\
\hline
$1$-block-norm MKL &   $  99.41 \pm 0.14$ \\
${4/3}$-block-norm MKL &  $  99.20 \pm 0.15$ \\
$2$-block-norm MKL     &   $ 99.25 \pm 0.15$ \\
$4$-block-norm    MKL   &  $ 99.14 \pm 0.16$ \\
$\infty$-block-norm  MKL    &  $ \mathbf{99.68} \pm \mathbf{0.09}$\\
\end{tabular}
\end{center}
\end{table}

Table \ref{tab:ids} shows the results for multiple kernel learning 
with various norms and elastic net parameters $\lambda$. 
The overall performance of all models is relatively high which is typical for intrusion detection applications.
 where very small false positive rates are crucial. 
The elastic net instantiations perform  relatively similar where $\mu=0.1$ is the most accurate one. It reaches
about the same level as $\ell_1$-block-norm MKL, which performs better than
the non-sparse $\ell_{4/3}$-norm MKL, the $\ell_{4}$-norm MKL, and the SVM with an unweighted-sum kernel.
Out of the block norm MKL versions---as already witnessed in the bioinformatics experiment---$\ell_\infty$-norm MKL gives the best predictor.

\section{Conclusion}\label{SEC-conclusion}

We presented a  framework for multiple kernel learning, that unifies several recent lines of research in that area. We phrased the seemingly different MKL variants as 
a single generalized optimization criterion and derived its dual. By plugging in an arbitrary convex loss function
many existing approaches can be recovered as instantiations of our model. We compared the different MKL variants in terms of their generalization performance 
by giving an concentration inequality for generalized MKL that matches the previous known bounds for $\ell_1$ and $\ell_{4/3}$ MKL.
We showed on artificial data how the optimal choice of an MKL model depends on the properties of the true underlying scenario. We compared several existing MKL instantiations on bioinformatics and network intrusion detection data.
Surprisingly, our empirical analysis shows that the recent uniformly non-sparse $\ell_\infty$ MKL of \cite{Nathetal09} outperforms its sparse and non-sparse competitors in both practical cases. It is up to future research to determine whether this empirical success also translates to other loss functions than hinge loss and other performance measures than the area under the ROC curve.

\small


\bibliographystyle{plain}

\end{document}